\documentclass[11pt]{article}
\usepackage{mtsummit2019}
\usepackage{times}
\usepackage{url}
\usepackage{latexsym}
\usepackage[small,bf]{caption} 
\usepackage{xcolor}
\usepackage{tikz}
\usepackage{graphicx}
\usepackage{enumitem}
\usetikzlibrary{shapes.geometric, arrows}
\usepackage{multirow}
\usepackage{placeins}
\usepackage{arydshln}

\tikzstyle{startstop} = [rectangle, rounded corners, minimum width=2cm, minimum height=0.5cm,text centered, draw=black, fill=white!30]
\tikzstyle{transl} = [rectangle, minimum width=2cm, minimum height=0.5cm,text centered, draw=black, fill=red!30]
\tikzstyle{arrow} = [thick,->,>=stealth]

\title{Lost in Translation: \\Loss and Decay of Linguistic Richness in Machine Translation}

 \author{Eva Vanmassenhove\\
   \And
   Dimitar Shterionov\\
   \\
   ADAPT, School of Computing,
   Dublin City University, Dublin, Ireland\\
   {\tt firstname.lastname@adaptcentre.ie} \\
  \And
   Andy Way\\
   }

\date{}

\newcommand{\citet}{~\newcite}
\newcommand{\citep}{~\cite}

\begin{document}
\maketitle
\begin{abstract}
This work presents an empirical approach to quantifying the loss of lexical richness in Machine Translation (MT) systems compared to Human Translation (HT). Our experiments show how current MT systems indeed fail to render the lexical diversity of human generated or translated text. The inability of MT systems to generate diverse outputs and its tendency to exacerbate already frequent patterns while ignoring less frequent ones, might be the underlying cause for, among others, the currently heavily debated issues related to gender biased output. Can we indeed, aside from biased data, talk about an algorithm that exacerbates seen biases? 

\end{abstract}

\section{Introduction}

\citet{berman2000translation} observed that the translation process consists of deformation processes, one of which he refers to as `quantitative impoverishment', a loss of lexical richness and diversity. Although mitigated by a human translator, this loss is to some extent inevitable as it is hard to respect the multitude of signifiers and constructions when translating one language into another. While \citet{berman2000translation} studied the decrease of lexical richness of human translations (HTs) from a theoretical point of view, \citet{Kruger2012} demonstrated using empirical methods that there is indeed a lexical loss when comparing translations to original texts. In the field of Machine Translation (MT), \citet{klebanov2013} showed that Statistical MT (SMT) suffers considerably more from lexical loss than HTs in a study focused on lexical tightness and text cohesion. We are not aware of any other research in this direction. 

\begin{figure}[h!]
\centering
\begin{tikzpicture}[node distance=2cm]
\node (sw) [startstop, yshift=-0.5cm] {uncountable};
\node (tw1) [startstop, right of=sw, xshift=2cm, yshift=-0.5cm] {ind\'enombrable};
\node (tw2) [startstop, right of=sw, xshift=2cm, yshift=0cm] {incalculable};
\node (tw3) [startstop, right of=sw, xshift=2cm, yshift=0.5cm] {innombrable};
\draw [arrow] (sw.east) -- (tw1.west);
\draw [arrow] (sw.east) -- (tw2.west);
\draw [arrow] (sw.east) -- (tw3.west);
\end{tikzpicture}
\caption{One-to-many relation between an English source word and some of its possible French translations} \label{fig:otm}
\end{figure}
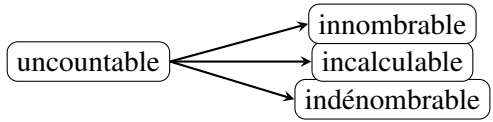

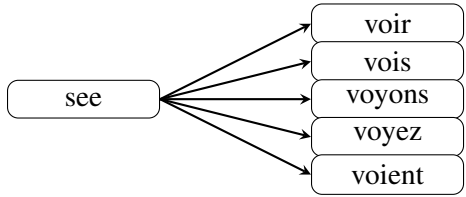
\begin{figure}[h!]
\centering
\begin{tikzpicture}[node distance=2cm]
\node (see) [startstop, yshift=-0.5cm] {see};
\node (voir) [startstop, right of=see, xshift=2cm, yshift=1cm] {voir};
\node (vois) [startstop, right of=see, xshift=2cm, yshift=0.5cm] {vois};
\node (voyons) [startstop, right of=see, xshift=2cm, yshift=0cm] {voyons};
\node (voyez) [startstop, right of=see, xshift=2cm, yshift=-0.5cm] {voyez};
\node (voient) [startstop, right of=see, xshift=2cm, yshift=-1cm] {voient};
\draw [arrow] (see.east) -- (voir.west);
\draw [arrow] (see.east) -- (vois.west);
\draw [arrow] (see.east) -- (voyons.west);
\draw [arrow] (see.east) -- (voyez.west);
\draw [arrow] (see.east) -- (voient.west);
\end{tikzpicture}
\caption{\textit{One-to-many} relation between English verb `see' and its conjugations in French} \label{fig:voir}
\end{figure}

\begin{figure}[h!]
\centering
\begin{tikzpicture}[node distance=2cm]
\node (smart) [startstop, yshift=-0.5cm] {smart};
\node (intelligente) [startstop, right of=smart, xshift=2cm, yshift=0.75cm] {intelligente };
\node (intelligent) [startstop, right of=smart, xshift=2cm, yshift=0.25cm] {intelligent  };
\node (intelligentes) [startstop, right of=smart, xshift=2cm, yshift=-0.25cm] {intelligentes};
\node (intelligents) [startstop, right of=smart, xshift=2cm, yshift=-0.75cm] {intelligents };
\draw [arrow] (smart.east) -- (intelligente.west);
\draw [arrow] (smart.east) -- (intelligent.west);
\draw [arrow] (smart.east) -- (intelligentes.west);
\draw [arrow] (smart.east) -- (intelligents.west);

\end{tikzpicture}
\caption{\textit{One-to-many} relation between English adjective `smart' and its male and female counterparts in French} \label{fig:genderb}
\end{figure}
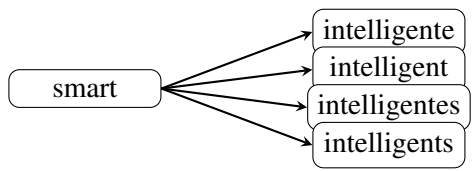

As generating accurate translations has been the main objective of current MT systems, maintaining lexical richness and creating diverse outputs has understandably not been a priority. Nevertheless, the issue of lexical loss in MT might at the same time be a symptom and a cause of a more serious issue underlying the current systems. The difference between a one-to-many relationship such as the one illustrated in Figure~\ref{fig:otm}, is very different from the one illustrated in Figure~\ref{fig:voir} or Figure~\ref{fig:genderb} from a (human) translator point of view. However, from a statistical point of view, they are not always clearly distinguishable. When presented with an ambiguous sentence, like `I am intelligent' or `See?' where there is little context to decide on a particular target variant of the same source word, it essentially boils down to the same thing: picking the translation that maximizes the probability over the entire sentence. As such, the loss of richness and diversity and the exacerbation of already frequent patterns might not simply be limited to the loss of (near) synonyms or rare words, but could also be the underlying cause of, for example, the inability of statistical MT systems to handle morphologically richer language correctly~\cite{Vanmassenhove2016,passban2018}, the already observed issues with gender bias~\cite{Vanmassenhove2018} in MT output or the difficulties of dealing with agglutinative languages~\cite{unanue2018}.

The inability of neural models to generate diverse output has already been observed for tasks involving language generation, where creating intrinsically diverse outputs is more of a necessity. However, from a translation point of view, the ability of MT systems to be (1) consistent and (2) learn and generalize well are --compared to previous MT systems-- the biggest asset of NMT. We however, hypothesize that this type of generalization might as well have serious drawbacks and that diversity, although not deemed a priority, is of importance for the field of MT as well. Overgeneralization over a seen input and the exacerbation of dominant forms might not only lead to a loss of lexical choice, but could also be the underlying cause of gender bias exacerbation. Although, in the context of gender, some researchers have already alluded to the existence of so-called `algorithmic bias' \cite{zhao2017men,Prates2018}, no empirical evidence has been provided so far.

With our empirical approach, comparing the lexical diversity of different MT systems and further analyzing the frequencies of words, we aim to shed some light on the relation between the loss of diversity and the exacerbation or loss of certain words. Thus, the first objective of our work is to verify how NMT compares to SMT and HT in terms of lexical richness or the loss thereof. The second objective is to quantify to what extent the different MT architectures favour translations that are more frequently observed in the training data.

The structure of the paper is the following: related work is described in Section~\ref{sec:relwork}; Our hypotheses are defined in details in Section~\ref{sec:hypothesis}; information on the data and the MT systems used in our experiments is provided in Section~\ref{sec:systems}; Section~\ref{sec:analysis} discusses the results of our experiments and finally, we conclude and provide some ideas for future work in Section~\ref{sec:conclusions}.

\section{Related Work}\label{sec:relwork}

In the field of linguistics,~\citet{berman2000translation} researched the so-called deforming tendencies that are inherent to the act of translation. Although these tendencies can be mitigated by the (human) translator, they are to a large extent inevitable. Quantitative impoverishment (or lexical loss), is one of the tendencies mentioned. \citet{Kruger2012} compared human-translated to comparable non-translated English texts and found the translations to be more simplified in terms of language use than the original writings.

In the field of MT, the concept of lexical loss/diversity and its importance is indirectly related to the research of~\citet{Wong2012} on cohesion. They illustrated the relevance of the under-use of linguistic devices (super-ordinates, meronyms, synonyms and near-synonyms) for SMT in terms of cohesion. More closely related to our work is the work of \citet{klebanov2013} who presented findings regarding the loss of associative texture by comparing original and back-translated texts, references and system translations and a set of different MT systems. Although the destruction of the underlying networks of signification might be, to some extent, unavoidable in any translation process, the work of~\citet{klebanov2013} shows that SMT specifically suffers from lexical loss, more than HT.

Lexical diversity or the loss thereof has also been used as a feature to estimate the quality of MT systems. \citet{Bentivogli2016neural} used lexical diversity, measured by using the type-token ratio (TTR), as an indicator of the size of vocabulary as well as the variety of subject matter in a text. Their experiments compared SMT to NMT and the results suggested that NMT is better able to cope with lexical diversity than SMT.

\section{Hypothesis}\label{sec:hypothesis} Data-driven statistical MT paradigms\footnote{Despite the fact that often phrase-based SMT is labeled as `statistical' and contrasted to `neural' MT or NMT, we ought to stress that both approaches are in fact \textit{statistical}.} are concerned with (i) identifying the most probable target words, phrases, or sub-word units given a source-language input sentence and the preceding decoded information, via the translation model, and (ii) chaining those words, phrases or sub-word units in a way that maximizes the likelihood of the generated sentence with respect to the grammatical and stylistic properties of the target language, via the language model. In NMT, where translation and language modeling are co-occurring in the decoder, it boils down to finding the most likely word at each time step.

Our hypothesis is that the inherent nature of data-driven MT systems to generalise over the training data has a quantitatively distinguishable negative impact on the word choice, expressed by favouring more frequent words and disregarding less frequent ones. We hypothesize that the most visible effect of such bias is to be found in the word frequencies and the disappearance (or `non-appearance') of scarce words. Apart from a general effect on lexical diversity, such behaviour might also lead to the disappearance or amplified use of certain morphological variants of the same word, accounting, for example, for the already observed over-use of male forms in ambiguous sentences, the preference for certain verb forms over other less frequent ones ($3^{rd}$ person $>$ $1^{st}$ person), or the difficulties of MT systems to appropriately handle morphologically richer target languages in general. 

Because NMT handles translation and language modelling (or alignment) jointly~\cite{Bahdanau2014,Vaswani2017}, which makes it harder to optimize compared to SMT, we further hypothesise that NMT is more susceptible to problems related to overgeneralisation. 

We present our experiments and analyses in Section~\ref{sec:exp} and Section~\ref{sec:analysis}.

\section{Empirical evaluation}\label{sec:exp}
To test our hypothesis we built three types of MT systems and analysed their output for two language pairs on Europarl~\cite{Koehn2005} data. The language pairs are English $\rightarrow$ French (EN-FR) and English $\rightarrow$ Spanish (EN-ES). We trained attentional RNN~\cite{Bahdanau2014}, Transformer~\cite{Vaswani2017} and Moses MT~\cite{Koehn2007} systems. To draw more general conclusions on the effects of bias propagation and loss of lexical richness, we assessed output from seen (during training) and unseen data.

\paragraph{Data} We used +/- 2M sentence pairs from the Europarl corpora for each of the language pairs. We randomised the order of the sentence pairs and split the data into train, test and development sets, filtering out empty lines. Details on the different datasets can be found in Table~\ref{tbl:data}. We chose to include large quantities of data in our test sets -- the unseen data -- in order to maximise the language variability and explore general tendencies.

\begin{table}[]
    \centering
    {\small \begin{tabular}{|c|c|c|c|}
    \hline
Language pair & Train & Test & Dev\\\hline
EN--FR & 1,467,489 & 499,487 & 7,723\\\hline
EN--ES & 1,472,203 & 459,633 & 5,734\\\hline
    \end{tabular}}
    \caption{Number of parallel sentences in the train, test and development splits for the language pairs we used.}
    \label{tbl:data}
\end{table}

\paragraph{MT systems}\label{sec:systems}
For each of the three MT architectures we first trained a standard MT system (the forward or FF system) on the original data. For the RNN and Transformer systems we used OpenNMT-py. The systems were trained for 150K steps, saving an intermediate model every 5000 steps. We scored the perplexity of each model on the development set and chose the one with the lowest perplexity as our best model, used later for translation. The options we used for the neural systems are as follows: 
\begin{itemize}[leftmargin=*]
    \item RNN: size: 512, RNN type: bidirectional LSTM, number of layers of the encoder and of the decoder: 4, attention type: mlp, dropout: 0.2, batch size: 128, learning optimizer: adam~\cite{Kingma2014} and learning rate: 0.0001.
    \item Transformer: number of layers: 6, size: 512, transformer\_ff: 2048, number of heads: 8, dropout: 0.1, batch size: 4096, batch type: tokens, learning optimizer adam with beta$_2 = 0.998$, learning rate: 2.
\end{itemize}
All neural systems have the learning rate decay enabled and their training is distributed over 4 nVidia 1080Ti GPUs. The selected settings for the RNN systems are optimal according to \cite{Britz2017}; for the Transformer we use the settings suggested by the OpenNMT community\footnote{\url{http://opennmt.net/OpenNMT-py/FAQ.html}} as the optimal ones that lead to quality on par with the original Transformer work~\cite{Vaswani2017}.

For the SMT systems we use Moses~\cite{Koehn2007} with default settings and a 5-gram language model with pruning of bigrams. Each system is further tuned with MERT~\cite{Och2003} until convergence or for a maximum of 25 iterations.

For the neural systems, we opted not to use sub-word units as is typically done for NMT. This is because we focus on the word frequencies in the translations and do not want any algorithm for splitting into sub-word units to add extra variability in our data. To construct the dictionaries we use all words in our training data. Table~\ref{tbl:vocabularies} (first two columns) shows the training vocabularies for the source and target sides.

\begin{table}[]
    \centering
    {\small 
     \setlength\tabcolsep{3pt} 
    \begin{tabular}{|c|c|c|}\hline
         Language pair & SRC & TRG \\\hline
EN--FR & 113,132 & 131,104 \\\hline
EN--ES & 113,692 & 168,195 \\\hline
    \end{tabular}}
    \caption{Training vocabularies for the English, French and Spanish data used for our models.}
    \label{tbl:vocabularies}
\end{table}

To assess how MT amplifies bias and loss of lexical richness, along with the original-data systems, we trained MT with backtranslated (BT) data, which is typically used to complement original data for MT training when the quantity of the original data is not sufficient for reaching high translation quality~\cite{Sennrich2015b,Poncelas2018}. 

\begin{figure}[h!]
\centering
\begin{tikzpicture}[node distance=4cm]

\node (in1) [startstop, label=north:{training data 1}] {FR - EN};
\node (tr1) [transl, label=north:{MT system 1: REV}, right of =in1] {FR $\Rightarrow$ EN*};
\node (in2) [startstop, label=north:{training data 2}, below of= in1, node distance=2cm] {EN* - FR};
\node (tr2) [transl,  label=north:{MT system 2: BACK}, right of=in2] {EN $\Rightarrow$ \textbf{FR*}};
\draw [arrow] (in1.east) -- (tr1.west);
\draw [arrow] (in2.east) -- (tr2.west);
\draw [arrow] (tr1.south) -- (in2.north);

\end{tikzpicture}
\caption{Back-translated data pipeline.} \label{fig:revback}
\end{figure}
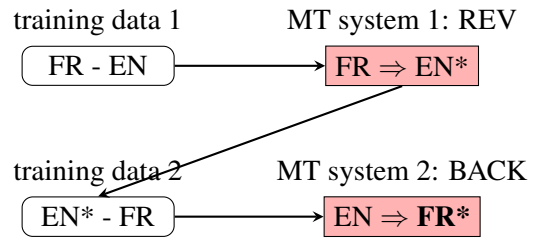

We first trained MT systems for the reverse language directions, i.e. for FR--EN and ES--EN. We used the same data sets, but reversed the associations of the source and the target with FR/ES $\rightarrow$ EN instead of EN $\rightarrow$ FR/ES. We then used these \textit{reversed} (\textit{REV} or \textit{rev}) systems to translate the training set: the same set used for training the FF systems and the REV systems. That is, we use a system trained on (say) FR--EN data to translate the same FR set into English (EN*). The aim is to see what is the impact of the underlying algorithms on the data in the most-favourable scenario; when the data has already been seen. With the translated English target data, we trained new systems for the EN*$\rightarrow$FR and EN*$\rightarrow$ES directions, where the source data was the backtranslated set. We refer to these systems with \textit{BACK} and use the suffix \textit{back} to denote them. We end up with what can be seen as a combination of back-translation and round-trip-translation. See Figure~\ref{fig:revback} for a visualization of the pipeline of systems.

For the REV and BACK systems we used the same settings as for the FF ones. However, at this stage, the source side of the training data is different and thus impacts the learnable vocabulary. Table~\ref{tbl:vocabularies_rev} presents the source-side vocabulary sizes for the RNN, SMT and Transformer systems. These are in practice the number of distinct words of the translations produced by the REV systems. Compared to Table~\ref{tbl:vocabularies}, this table clearly shows how source and target vocabularies are comparable in the original datasets, but translating the same original English dataset with the neural REV systems (RNN and Transformer) results in a huge drop in vocabulary size; with the SMT REV systems the decrease is still significant, but not as profound as in the former cases.


\begin{table}[]
    \centering
    {\small 
     \setlength\tabcolsep{2.2pt} 
    \begin{tabular}{|c|c|c|c|c|c|c|}\hline
        Lang. & \multicolumn{3}{c|}{EN*} & \multicolumn{3}{c|}{FR*/ES*}\\\cline{2-7}
         pair & RNN & SMT & Trans. & RNN & SMT & Trans. \\\hline
EN--FR & 28,742 & 106,441 & 40,321 & 36,991 & 123,770 & 42,309\\\hline
EN--ES & 27,349 & 118,362 & 40,629 & 39,805 & 138,193 & 44,545\\\hline
    \end{tabular}}
    \caption{Vocabularies of the English translation from the REV systems, used as source for the BACK systems and the French/Spanish output from the BACK systems.}
    \label{tbl:vocabularies_rev}
\end{table}
In Table~\ref{tbl:bleu} we present automatic evaluation scores -- BLEU~\cite{Papineni2002} and TER~\cite{Snover2006} -- for the 12 analysed systems. For completeness we present BLEU and TER for the REV systems in Table~\ref{tbl:bleu_rev}, although we do not consider them in our analysis. For the test set we performed a statistical significance test using the \emph{multeval} tool~\cite{Clark2011}. For $p < 0.05$ all results in Table~\ref{tbl:bleu} are statistically significant.

In what follows we use the following denotations to indicate the system we refer to: \verb|{src}-{trg}-{system}-{dir}|, where \verb|{src}| indicates the source language `en', that is English, \verb|{trg}| indicates the target languag -- `fr' for French and `es' for Spanish -- and the system is one of `HT' for human translation, `smt' for SMT, `rnn' for the RNN models and `trans' for the Transformer models; \verb|{dir}| is one of `ff' to indicate that the system is the forward, trained on the original data, `back' to indicate that the system is trained with back-translated data or `rev' to denote that it is the reverse system, trained after swapping source and target (the human translation has no \verb|dir| index).

\begin{table}[]
    \centering
    {\small 
     \setlength\tabcolsep{3pt} \begin{tabular}{|l|c|c|c|c|}\hline
        \multicolumn{1}{|c|}{System} & \multicolumn{2}{c|}{Dev set} & \multicolumn{2}{c|}{Test set}\\\cline{2-5}
        \multicolumn{1}{|c|}{reference}  & BLEU$\uparrow$ & TER$\downarrow$ & BLEU$\uparrow$ & TER$\downarrow$\\\hline
en-fr-rnn-ff & 33.7 & 50.7 & 33.8 & 51.0\\\hline
en-fr-smt-ff & 35.9 & 50.4 & 35.7 & 50.7\\\hline
en-fr-trans-ff & 35.9 & \textbf{49.5} & 36.0 & \textbf{49.4}\\\hline
en-fr-rnn-back & 32.8 & 52.1 & 33.0 & 52.1\\\hline
en-fr-smt-back & 35.2 & 51.0 & 35.0 & 51.3\\\hline
en-fr-trans-back & \textbf{36.3} & 49.8 & \textbf{36.3} & 49.9\\\hline\hline
en-es-rnn-ff & 37.4 & 45.3 & 37.9 & 45.3\\\hline
en-es-smt-ff & 38.5 & 45.8 & 38.6 & 45.9\\\hline
en-es-trans-ff & \textbf{39.4} & \textbf{44.5} & \textbf{39.5} & \textbf{44.5}\\\hline
en-es-rnn-back & 36.0 & 47.0 & 36.3 & 47.0\\\hline
en-es-smt-back & 38.0 & 46.5 & 38.0 & 46.5\\\hline
en-es-trans-back & \textbf{39.4} & 45.2 & 39.3 & 45.5\\\hline
    \end{tabular}}
    \caption{Automatic evaluation scores (BLEU, TER) for all MT systems.}
    \label{tbl:bleu}
\end{table}

\begin{table}[ht]
    \centering
     {\small 
     \setlength\tabcolsep{3pt} \begin{tabular}{|l|c|c|}\hline
        \multicolumn{1}{|c|}{System reference} & BLEU$\uparrow$ & TER$\downarrow$ \\\hline
        en-fr-rnn-rev & 33.3 & 50.2\\\hline
en-fr-smt-rev & 36.5 & 47.1\\\hline
en-fr-trans-rev & \textbf{36.8} & \textbf{46.8}\\\hline
en-es-rnn-rev & 37.8 & 45.0\\\hline
en-es-smt-rev & 39.2 & 44.0\\\hline
en-es-trans-rev & \textbf{40.4} & \textbf{42.7}\\\hline
    \end{tabular}}
    \caption{Automatic evaluation scores (BLEU and TER) for the REV systems.}
    \label{tbl:bleu_rev}
\end{table}
\paragraph{Evaluated output}
In total we trained 18 MT systems. To assess the validity of our hypothesis and to provide a quantitative analysis of the investigated phenomena, we use the outputs from the FF and the BACK systems; the REV systems are used just to generate the backtranslated data.

\section{Analysis}\label{sec:analysis}
In the analysis we compare word frequencies of the original target data to the translation output of the forward (FF) and backward (BACK) MT systems. We investigate two scenarios: (i) \textit{seen} and (ii) \textit{unseen} data. For (i) we translate the original source side of the training set (i.e. the English sentences) with the FF and with the BACK systems. The reason behind performing this kind of test is that since the MT system has seen this data during training, any loss of lexical richness and/or bias exacerbation are due to the inherent workings of the systems. That is, the observed differences between lexical diversity on seen data can only be attributed to the algorithm itself. For (ii) we are evaluating the lexical diversity on the (unseen) test set. This evaluation scenario is the one that gives us an indication of the overall lexical diversity of the translations produced by MT systems as compared to the data they were trained on. 

\paragraph{Language diversity score}
Lexical diversity (LD) refers to the amount or range of different words that are used in a text. The greater that range, the higher the diversity. Although LD has many applications (neuropathology, data mining, language acquisition), coming up with a robust index to quantify it has proven to be a difficult task. A comparison between different measures of LD~\cite{mccarthy2010} concluded by saying that, although there is no consensus yet, LD can be assessed in different ways, with each measurement having its own assets and drawbacks. Therefore, we evaluated LD by using four different widely used metrics: type/token ratio (TTR)~\cite{Templin1975certain}, Yule's K (in practice, we use the reverse Yule's I)~\cite{Yule1944}, and the measure of textual lexical diversity (MTLD)~\cite{Mccarthy2005assessment}. 

The easiest lexical richness metric is TTR. TTR is the ratio of the types, i.e. total number of \textit{different} words in a text to its tokens, i.e. the total number of words. A high/low TTR indicates a high/low degree of lexical diversity. While TTR is one of the most widely used metrics, it has some drawbacks linked to the assumption of a linear relation between the types and the tokens. Because of that, TTR is only valid when comparing texts of a similar size, as it decreases when texts become longer due to repetitions of words \cite{brezina2018}. 

Yule's characteristic constant, or Yule's K, is a probability model of the changes that take place in the lexical frequency spectrum of a text as the text becomes longer. Yule's K and its reverse Yule's I are considered to be more immune to fluctuations related to text length than TTR~\cite{oakes2012}. 

Another metric used to study lexical richness and diversity is MTLD. The difference with the two previous methods is that MTLD is evaluated sequentially as the mean length of sequential word strings in a text that maintain a given TTR value \cite{Mccarthy2005assessment}. A more recent study by \citet{mccarthy2010} shows that MTLD is the most robust with respect to text length.





Our metrics are presented in Table~\ref{tbl:results_train} and Table~\ref{tbl:results_test}. Higher/lower scores indicate higher/lower lexical richness. Table~\ref{tbl:results_train} shows the metrics for the human and the machine translations of the training set, i.e. the seen data, and Table~\ref{tbl:results_test} shows the scores for the human (HT) and the machine translations of the test sets, i.e. the unseen data. Due to the large number of output words, e.g. the rnn-ff translation of the EN--FR test set contains 14 561 653 words, and the low vocabulary size relative to the total number of words, our TTR scores are quite low. For readability and for ease of comparison we present these scores multiplied by a factor of 1000. We tested pairwise statistical significance through bootstrap sampling following~\cite{Koehn2004-statistical-significance}. The scores for all MT variants are significantly different from the HT variant. 

\begin{table}[ht]
     \centering
    {\small 
      \begin{tabular}{|l|c|c|c|}\hline
         Translation & Yule’s I & TTR & MTLD\\
          &  & * 1000 & \\\hline\hline
en-fr-HT & \textit{9.2793} & \textit{2.9277} & \textit{127.1766} \\\hline\hline
en-fr-rnn-ff & 0.7107 & 0.8656 & 109.4506 \\\hline
en-fr-smt-ff & 6.7492 & 2.6442 & 118.1239 \\\hline
en-fr-trans-ff & 1.1768 & 1.0925 & 120.5179\\\hline
en-fr-rnn-back & 0.7587 & 0.8776 & 116.8942\\\hline
en-fr-smt-back & \textbf{7.8738} & \textbf{2.7496} & 120.9909\\\hline
en-fr-trans-back & 1.0325 & 1.0172 & \textbf{121.5801}\\\hline\hline
en-es-HT & \textit{12.3065} & \textit{3.7037} & \textit{99.0850}\\\hline\hline
en-es-rnn-ff & 0.6298 & 0.9394 & 89.3562\\\hline
en-es-smt-ff & 7.3249 & 3.1170 & 95.1146\\\hline
en-es-trans-ff & 1.0022 & 1.1581 & \textbf{96.2113}\\\hline
en-es-rnn-back & 0.7355 & 0.9829 & 95.7198\\\hline
en-es-smt-back & \textbf{8.1325} & \textbf{3.2166} & 95.1479\\\hline
en-es-trans-back & 0.9162 & 1.1014 & 95.0886\\\hline
    \end{tabular}}
    \caption{Lexical richness metrics (Train set).}
    \label{tbl:results_train}
\end{table}

\begin{table}[ht]
     \centering
    {\small 
     \begin{tabular}{|l|c|c|c|}\hline
         Translation & Yule’s I & TTR & MTLD\\
          &  & * 1000 & \\\hline\hline
en-fr-HT & 33.6709 & 5.7022 & 124.1889\\\hline\hline
en-fr-rnn-ff & 4.4766 & 2.1969 & 106.1370\\\hline
en-fr-smt-ff & 21.1230 & 4.8034 & 113.9262\\\hline
en-fr-trans-ff & 6.5352 & 2.5957 & 118.9642\\\hline
en-fr-rnn-back & 5.1490 & 2.3092 & 112.9991\\\hline
en-fr-smt-back & \textbf{25.7705} & \textbf{5.1254} & 117.6979\\\hline
en-fr-trans-back & 6.7921 & 2.6287 & \textbf{119.1729}\\\hline\hline
en-es-HT & 48.2366 & 7.6151 & 97.0591\\\hline\hline
en-es-rnn-ff & 4.7988 & 2.6250 & 85.4589\\\hline
en-es-smt-ff & 24.6771 & 5.9171 & 92.6397\\\hline
en-es-trans-ff & 6.7967 & 3.0432 & \textbf{94.4709}\\\hline
en-es-rnn-back & 6.0098 & 2.8357 & 92.4704\\\hline
en-es-smt-back & \textbf{28.0153} & \textbf{6.1887} & 92.3310\\\hline
en-es-trans-back & 7.3824 & 3.1483 & 92.8928\\\hline
    \end{tabular}}
    \caption{Lexical richness metrics (Test set).}
    \label{tbl:results_test}
\end{table}
\paragraph{Word frequencies and bias} 
In order to prove/disprove our hypothesis, along with investigating lexical richness, we aim to investigate to what extent MT systems propagate bias in the output. This we assess by whether more/less frequent words in the human translation have higher/lower frequency in the MT output (see Section~\ref{sec:hypothesis}). As soon as we started training the BACK systems, the first thing we observed was the reduced vocabularies from the FF systems. The loss of certain words (in the case of unknown words, the RNN and Transformer systems would generate the \verb|<unk>| token) already suggests biased MT. Comparing Table~\ref{tbl:vocabularies} and Table~\ref{tbl:vocabularies_rev}, we see that a lot of words are not accounted for in all systems, but that the RNN and Transformer models suffer the most. We believe this is due to the fact that NMT systems' advantage over more traditional systems, namely its ability to generalize and learn over the entire sentence, has a negative affect on lexical diversity, particularly for the least frequent words. 

Due to the differences in vocabularies and sentence lengths of the generated translations, in order to conduct a realistic comparison of the frequencies we applied 3 post-processing steps on the collected data: (i) we accounted for sentence variability by normalizing the frequency of each word (in the HT or the MT output) by the length of sentences in which it appears, (ii) we normalized the frequency of each word (in the HT or the MT output) by the accumulated frequency, reducing each frequency to a probability, and (iii) to account for the missing words in the MT output we counted words with zero frequencies separately. In addition, we need to make a distinction between frequent and non-frequent words. While this is a hard task in itself, here we commit to the average normalized word frequency of the human translation.

Once we applied the aforementioned post-processing we compactly represent our data in six classes: 
\begin{itemize}
    \item \textit{Frequency increase of frequent words}: for a frequent word in the HT, its frequency in the MT is higher.  We denote this class using `\verb|+ +|' symbol combination. This class also indicates positive bias exacerbation.
    \item \textit{Frequency decrease of frequent words}: for a frequent word in the HT, its frequency in the MT is lower (but not zero).  We denote this class using `\verb|+ -|' symbol combination. 
    \item \textit{Frequency increase of non-frequent words}: for a non-frequent word in the HT, its frequency in the MT is higher.  We denote this class using `\verb|- +|' symbol combination. 
    \item \textit{Frequency decrease of non-frequent words}: for a non-frequent word in the HT, its frequency in the MT is lower (but not zero).  We denote this class using `\verb|- -|' symbol combination. This class indicates negative bias exacerbation.
    \item \textit{Zero frequency of frequent words}: a frequent word in the HT, does not appear in the MT.  We denote this class using `\verb|+ 0|' symbol combination. 
    \item \textit{Zero frequency of non-frequent words}: a non-frequent word in the HT, does not appear in the MT.  We denote this class using `\verb|- 0|' symbol combination. This class indicates negative bias exacerbation.
\end{itemize}

For each of these classes we count the (normalized) number of words, and we accumulate the absolute value of the differences for each of these cases. We present our results for the training data in Table~\ref{tbl:ex_freq_train}, Table~\ref{tbl:acc_freq_train} and for the test data -- in Table~\ref{tbl:ex_freq_test}, Table~\ref{tbl:acc_freq_test}. The numbers in Table~\ref{tbl:ex_freq_train} and Table~\ref{tbl:ex_freq_test} can be interpreted as the amount of translated words with higher, lower or zero frequency compared to the human translation.\footnote{Note that these numbers are normalized for fair comparison.} The numbers in Table~\ref{tbl:acc_freq_train} and Table~\ref{tbl:acc_freq_test} quantify the differences between frequencies; they indicate the amount of increase or decrease in the frequencies presented by an MT system as compared to the HT. To derive information from these numbers, one should compare the '\verb|+ +|' to `\verb|+ -|' and `\verb|- +|' to `\verb|- -|' and '\verb|+ 0|' to `\verb|- 0|'.

\begin{table}[ht]
    \centering
    {\small 
     \setlength\tabcolsep{2.1pt} \begin{tabular}{|l|c|c|c|c|c|c|}\hline
         System & + + & + - & - + & - - & + 0 & - 0\\\hline
en-fr-rnn-ff & 3710 & 3023 & 10157 & 18683 & 10 & 95519\\\hline
en-fr-smt-ff & 3362 & 3381 & 32577 & 46714 & 0 & 45068\\\hline
en-fr-trans-ff & 3839 & 2901 & 12398 & 24403 & 3 & 87558\\\hline
en-fr-rnn-back & 3356 & 3372 & 13009 & 17253 & 15 & 94097\\\hline
en-fr-smt-back & 3246 & 3496 & 34111 & 43472 & 1 & 46776\\\hline
en-fr-trans-back & 3482 & 3254 & 14610 & 20962 & 7 & 88787\\\hline\hline
en-es-rnn-ff & 4667 & 3532 & 9929 & 19149 & 41 & 130875\\\hline
en-es-smt-ff & 4276 & 3963 & 39817 & 56169 & 1 & 63967\\\hline
en-es-trans-ff & 4626 & 3601 & 11379 & 25698 & 13 & 122876\\\hline
en-es-rnn-back & 4265 & 3951 & 13716 & 17872 & 24 & 128365\\\hline
en-es-smt-back & 4006 & 4233 & 39636 & 51831 & 1 & 68486\\\hline
en-es-trans-back & 4288 & 3929 & 14295 & 22032 & 23 & 123626\\\hline
    \end{tabular}}
    \caption{Frequency exacerbation and decay count (Train set)}
    \label{tbl:ex_freq_train}
\end{table}
\begin{table}[ht]
    \centering
    {\small 
     \setlength\tabcolsep{1.2pt} \begin{tabular}{|l|c|c|c|c|c|c|}\hline
         System & + + & + - & - + & - - & + 0 & - 0\\\hline
en-fr-rnn-ff & 2917 & 2335 & 10653 & 15400 & 11 & 57623\\\hline
en-fr-smt-ff & 2652 & 2610 & 20587 & 26949 & 1 & 36140\\\hline
en-fr-trans-ff & 2997 & 2264 & 12537 & 17430 & 2 & 53709\\\hline
en-fr-rnn-back & 2642 & 2610 & 13513 & 14963 & 11 & 55200\\\hline
en-fr-smt-back & 2577 & 2684 & 22604 & 26608 & 2 & 34464\\\hline
en-fr-trans-back & 2701 & 2554 & 14932 & 17101 & 8 & 51643\\\hline\hline
en-es-rnn-ff & 3541 & 2669 & 10636 & 16425 & 27 & 75113\\\hline
en-es-smt-ff & 3252 & 2982 & 23389 & 29057 & 3 & 49728\\\hline
en-es-trans-ff & 3508 & 2716 & 12069 & 19046 & 13 & 71059\\\hline
en-es-rnn-back & 3241 & 2971 & 14394 & 15847 & 25 & 71933\\\hline
en-es-smt-back & 3163 & 3072 & 24547 & 28389 & 2 & 49238\\\hline
en-es-trans-back & 3256 & 2967 & 15160 & 18606 & 14 & 68408\\\hline
    \end{tabular}}
    \caption{Frequency exacerbation and decay count (Test set)}
    \label{tbl:ex_freq_test}
\end{table}
\begin{table}[ht]
    \centering
    {\small 
     \setlength\tabcolsep{1.2pt} \begin{tabular}{|l|c|c|c|c|c|c|}\hline
         System & + + & + - & - + & - - & + 0 & - 0\\\hline
en-fr-rnn-ff & 840.76 & 687.16 & 46.36 & 115.27 & 1.47 & 83.22\\\hline
en-fr-smt-ff & 664.86 & 555.60 & 31.17 & 119.64 & 0.00 & 20.79\\\hline
en-fr-trans-ff & 663.00 & 552.74 & 49.98 & 108.63 & 0.40 & 51.20\\\hline
en-fr-rnn-back & 770.72 & 680.73 & 83.68 & 96.68 & 2.19 & 74.81\\\hline
en-fr-smt-back & 620.67 & 525.26 & 40.36 & 112.35 & 0.13 & 23.29\\\hline
en-fr-trans-back & 639.69 & 568.68 & 75.88 & 90.25 & 1.05 & 55.58\\\hline\hline
en-es-rnn-ff & 733.44 & 535.15 & 42.54 & 117.47 & 4.93 & 118.43\\\hline
en-es-smt-ff & 547.86 & 423.87 & 33.22 & 129.73 & 0.12 & 27.35\\\hline
en-es-trans-ff & 587.22 & 436.02 & 47.61 & 119.98 & 1.37 & 77.46\\\hline
en-es-rnn-back & 677.23 & 564.31 & 94.47 & 101.57 & 2.92 & 102.90\\\hline
en-es-smt-back & 561.03 & 438.09 & 44.31 & 133.35 & 0.12 & 33.78\\\hline
en-es-trans-back & 548.37 & 438.33 & 72.27 & 98.11 & 2.33 & 81.87\\\hline
    \end{tabular}}
    \caption{Accumulated frequency differences (Train set)}
    \label{tbl:acc_freq_train}
\end{table}
\begin{table}[ht]
    \centering
    {\small 
     \setlength\tabcolsep{1.2pt} \begin{tabular}{|l|c|c|c|c|c|c|}\hline
         System & + + & + - & - + & - - & + 0 & - 0\\\hline
en-fr-rnn-ff & 827.07 & 655.81 & 68.84 & 133.21 & 2.48 & 104.41\\\hline
en-fr-smt-ff & 790.41 & 640.60 & 60.98 & 156.94 & 0.13 & 53.71\\\hline
en-fr-trans-ff & 662.76 & 533.83 & 73.15 & 123.07 & 0.31 & 78.70\\\hline
en-fr-rnn-back & 751.49 & 655.35 & 112.32 & 114.16 & 2.28 & 92.01\\\hline
en-fr-smt-back & 679.17 & 551.88 & 64.50 & 142.50 & 0.34 & 48.96\\\hline
en-fr-trans-back & 625.59 & 548.18 & 104.26 & 107.39 & 1.41 & 72.88\\\hline\hline
en-es-rnn-ff & 726.16 & 509.28 & 67.76 & 134.45 & 4.16 & 146.04\\\hline
en-es-smt-ff & 679.08 & 503.57 & 70.86 & 169.33 & 0.38 & 76.67\\\hline
en-es-trans-ff & 592.32 & 414.37 & 73.00 & 134.59 & 1.84 & 114.52\\\hline
en-es-rnn-back & 653.89 & 533.03 & 128.86 & 119.04 & 4.22 & 126.46\\\hline
en-es-smt-back & 630.86 & 462.82 & 74.19 & 165.11 & 0.31 & 76.81\\\hline
en-es-trans-back & 538.03 & 415.49 & 103.32 & 118.89 & 2.40 & 104.57\\\hline
    \end{tabular}}
    \caption{Accumulated frequency differences (Test set)}
    \label{tbl:acc_freq_test}
\end{table}

\paragraph{Remarks on automatic evaluation}
The summary of our results allows us to draw the following conclusions:
\begin{enumerate}[leftmargin=*]
    \item \textit{Lexical richness} All metrics and results presented in Table~\ref{tbl:results_train} and Table~\ref{tbl:results_test} and for both language pairs indicate that neither of the MT systems reaches the lexical richness of the HT. While SMT systems (for both language pairs) retain more language richness according to two out of the three metrics (Yule's I and TTR) than the neural methods, the MTLD scores indicate that the Transformer systems lead to translations of higher lexical richness. This we may account for the different numbers of distinct words produced by SMT and neural systems, which may be favoured by Yule's I and TTR. However, consistently, the worst systems are the RNN ones. 
    \item \textit{Automatic quality evaluation vs. lexical richness}: The results in Table~\ref{tbl:bleu} show that the Transformer systems perform best. The only lexical richness metric that corroborates the BLEU and TER scores is MTLD. This observation can act as a future research direction for integrating or improving quality evaluation metrics of MT to accommodate for lexical richness by possibly adopting features from MTLD.
    \item \textit{Bias} To understand how the inherent probabilistic nature of PB-SMT and NMT systems exacerbates (or not) the bias, we rely on the result in Table~\ref{tbl:ex_freq_train}, Table~\ref{tbl:ex_freq_test}, Table~\ref{tbl:acc_freq_train} and Table~\ref{tbl:acc_freq_test}. More precisely, we focus on the comparison between `\verb|[+ +]|' and `\verb|[+ -]|', and the `\verb|[- +]|' and `\verb|[- -]|' classes as well as the values in the `\verb|[+ 0]|' and `\verb|[- 0]|' classes. One could simplify the analysis by joining the latter two classes together with `\verb|[+ -]|' and `\verb|[- -]|'. However, their independent analysis carries more important information. Precisely, we see that all of the systems lose less frequent words, indicated by the low numbers for the `\verb|[+ 0]|' class for both the training and the test set translations. Second, not all MT systems produce more words with higher frequencies (for the Train set: en-fr-PB-SMT-ff with 3362 vs 3381, en-fr-PB-SMT-back with 3246 vs 3496 and en-es-PB-SMT-back with 4006 vs 4233; for the test set: en-fr-PB-SMT-back with 2577 vs 2684), but the accumulative normalized frequency for such words is higher than that of the HT. The accumulated frequency differences indicate that MT systems are indeed biased towards these more frequent words. This observation, together with the fact that all MT systems suffer from loss of less frequent words, further supports our hypothesis that MT systems target learning the more frequent words and disregard the less frequent ones.
    \item \textit{Seen and unseen data}
    We divided our experiments over seen and unseen data. From the perspective of lexical richness we see the same trends in both cases, although a slight decrease can be observed for the unseen test set (measured by the MTLD metric). With regards to the word frequencies comparing `\verb|+ +|' and `\verb|+ -|' classes in Table~\ref{tbl:acc_freq_train} and Table~\ref{tbl:acc_freq_test} we see similar trends. Furthermore, more words are lost altogether when translating the unseen test set. 
\end{enumerate}

It should be stressed that in this work we looked at the frequency of words, and as such the RNN and Transformer models we trained are not optimized according to state-of-the-art settings. In particular, no BPE is used to account for out-of-vocabulary problems, and the vocabularies have not been restricted prior to training (typically the vocabulary of an NMT system consists of the K, e.g. $50$k most frequent words/tokens).

Another observation that we ought to note is that the BACK systems score quite high not only based on word frequencies and lexical richness metrics, but also based on the evaluation metrics presented in Table~\ref{tbl:bleu}. We assume this is due to the fact that the simplified source (translated by the REV systems) changes the complexity of the learned association. We plan to further investigate these systems. 

\paragraph{Semi-manual evaluation}
To obtain a more concrete image of the observed bias exacerbation by MT, we looked into the translations of 15 random English words: `picture', `create', `states', `happen', `genuine', `successful', `also', `reasons', `membership', `encourage', `selling', `site', `vibrant', `still' and `event'. This evaluation does not have the intention to be exhaustive, as the general tendencies of the systems have already been discussed in the previous sections. However, looking into some actual translations produced by the systems does further clarify the exacerbation effect of the learning algorithm. 

\begin{figure}[ht]
    \centering
        \includegraphics[width=0.45\textwidth]{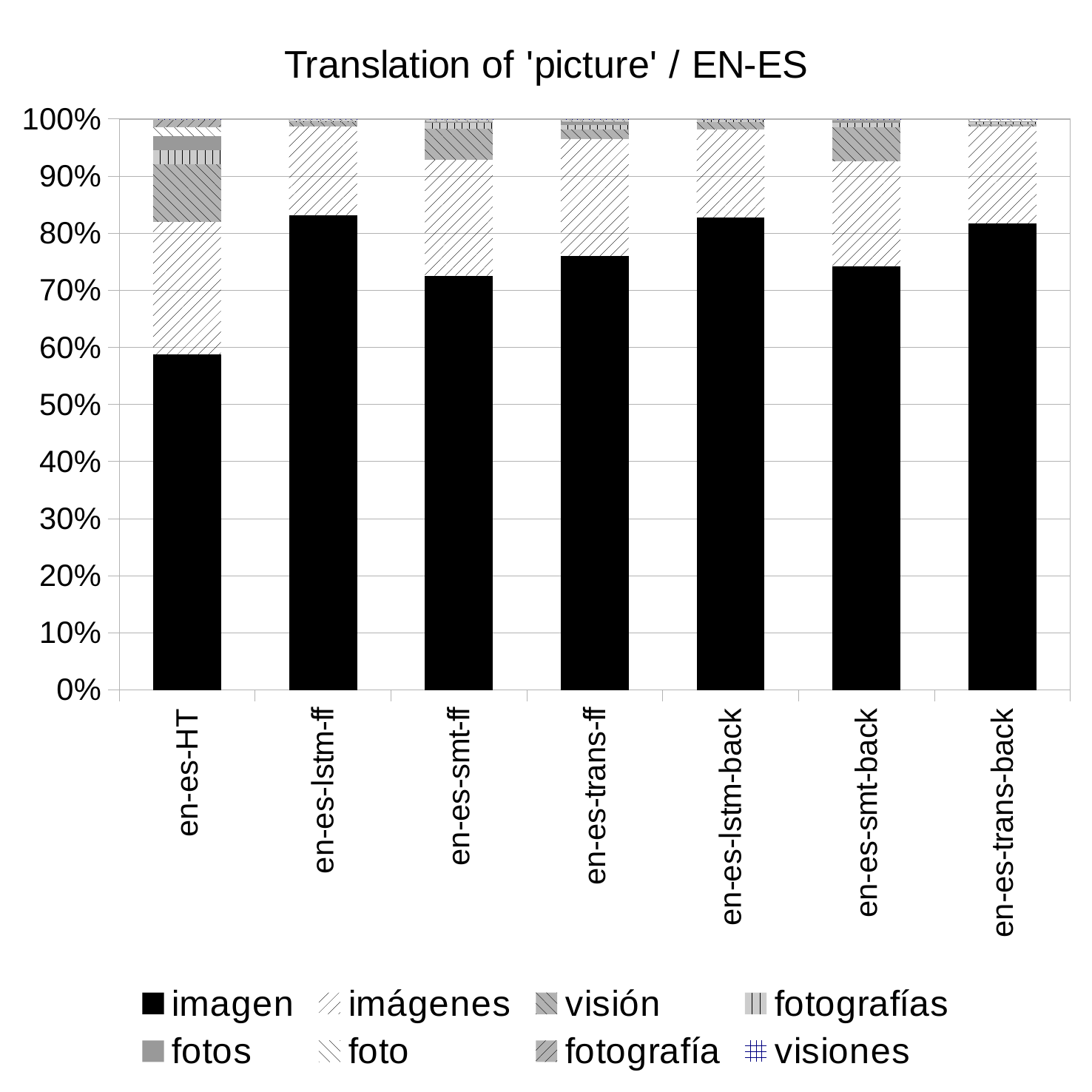}
        \includegraphics[width=0.45\textwidth]{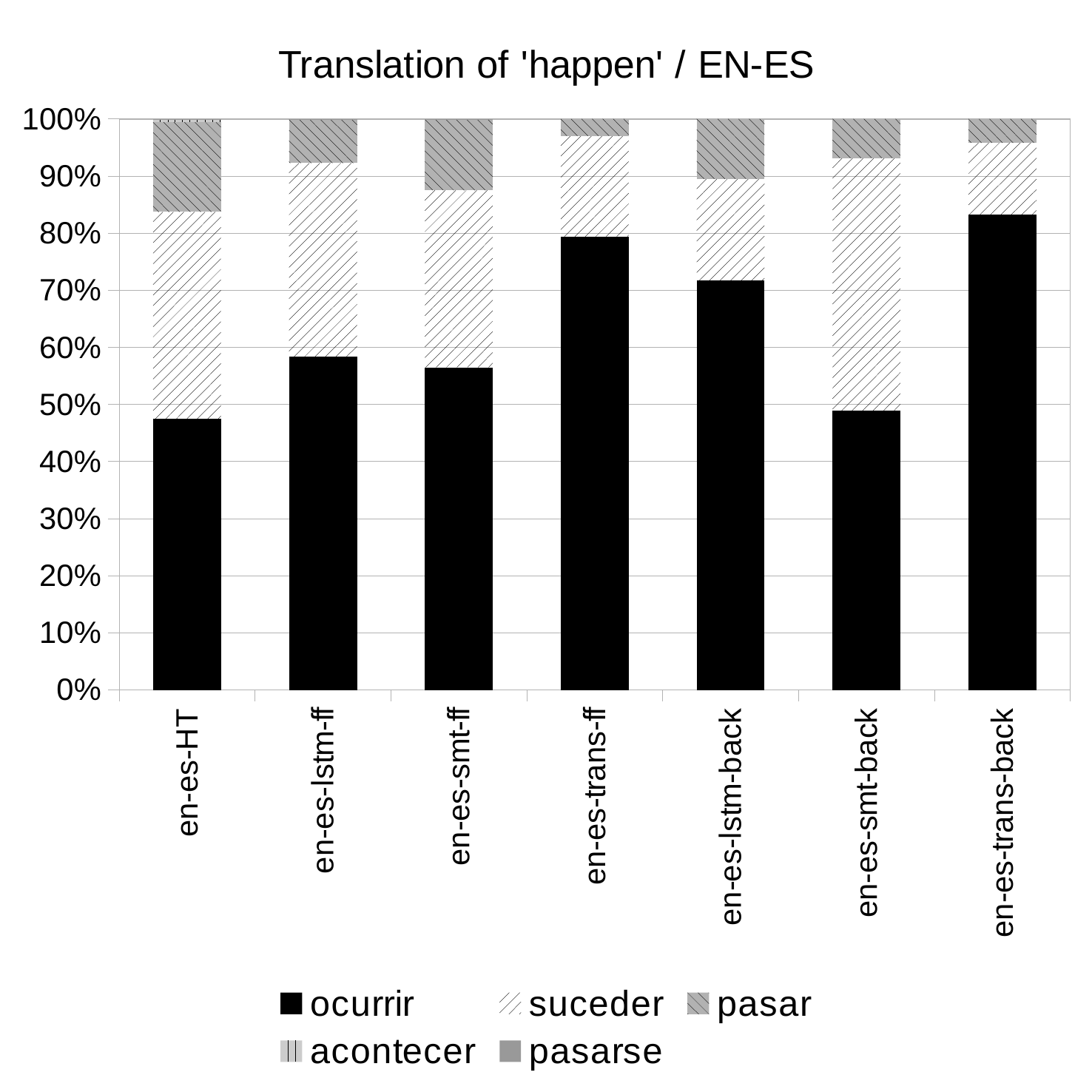}
    \caption{Relative frequencies of the Spanish translations of the English words `picture' and `happen'.}
    \label{fig:translations_es}
\end{figure}

Let us first look at the Spanish translations of the English word `picture', presented in Figure~\ref{fig:translations_es}. The original data shows quite a lot of diversity as `picture' can be translated into among others `imagen', `im\'agenes', `visi\'on', `foto',`fotograf\'ias' and `fotos'. However, when we look at the output of the EN--ES MT systems, we see that all of them use the most frequent translation --`imagen'-- even more frequently than in the original data. This comes at the expense of the other translation variants. Although the second most frequent translation (`im\'agenes') is still frequent, all others show a decrease and the least frequent ones disappear entirely.

Similar, though slightly different patterns are observed for the translations of the other words we examine. Also presented in Figure~\ref{fig:translations_es} are the translations of the English verb `happen' into the Spanish verbs `ocurrir', `suceder',`pasar', `acontecer' and `pasarse' and the English conector `also' into `tambi\'en', `adem\'as' and `igualmente'. Again, the graphs show how the most frequent translation(s) gain in relative frequency at the cost of less frequent options.

\section{Conclusions and Future Work}\label{sec:conclusions}
This work investigates bias exacerbation and loss of lexical richness through the process of MT. We analyse the problem of loss of lexical richness using a number of LD metrics on the output of 12 different MT systems: SMT, RNN and Transformer models for EN--FR and EN--ES with original and back-translated data. 

Via our experiments and their subsequent analysis, we observe that the process of MT causes a general loss in terms of lexical diversity and richness when compared to human-generated text. This confirms our first hypothesis. Furthermore, we investigate how this loss comes about and whether it is indeed the case that the more frequent words observed in the input occur even more in the output, negatively affecting the frequency of less seen events or words by causing them to become even rarer events or causing them to disappear altogether. Our analysis shows that MT paradigms indeed increase/decrease the frequencies of more/less frequent words to such extend that a very large amount of words are completely `lost in translation'. We believe, this demonstrates indeed that current systems overgeneralize and thus, we deem it appropriate to speak of a form of algorithmic bias. 

Overall, the RNNs systems are among the worst performing in terms of LD, although we do need to take into account that, for the sake of comparison, we did not use BPE, which might gave the neural models a disadvantage compared to the SMT systems. While Transformer models are the best ones according to the evaluation metrics, SMT seems to retain the most lexical richness according to the LD metrics we used (TTR, Yule's I and MTLD).

As research on language generation has already accounted for the lack of diverse outputs, in the future, we aim to lock into potential solutions to overgeneralization of current trnaslation models. However, allowing for a certain degree of randomness while maintaining a strong learning (and thus generalizing) ability is a very complex and potentially contradictory task. 




 \section*{Acknowledgements}\label{sec:ack}

This work has been supported by the Dublin City University Faculty of Engineering \& Computing under the Daniel O'Hare Research Scholarship scheme and by the ADAPT Centre for Digital Content Technology, which is funded under the SFI Research  Centres  Programme (Grant  13/RC/2106).\\

\bibliography{mtsummit2019}
\bibliographystyle{mtsummit2019}

\end{document}